\documentclass{article}
\usepackage{ijcai26}
\usepackage[T1]{fontenc}
\usepackage[utf8]{inputenc}
\usepackage{times}
\usepackage{xcolor}
\usepackage{soul}
\usepackage{url}
\usepackage[hidelinks]{hyperref}
\usepackage{graphicx}
\usepackage{subcaption}
\usepackage{amsmath,amssymb,amsthm}
\usepackage{booktabs}
\usepackage{multirow}
\usepackage[numbers]{natbib}
\usepackage{braket}
\usepackage{physics}
\usepackage{tikz}
\usepackage{quantikz}
\usepackage{float}
\usepackage{cleveref}

\title{Quantum-Inspired Hybrid Neural Networks for Neural Decoding:\\
A Controlled Ablation Study of Learnable Quantum Sidecar Integration}

\author{
Diana Legziel Levy$^1$
\and Menachem Finkelstein$^1$
\and Peter Chin$^2$
\and Eilon Vaadia$^3$
\and Sarel Cohen$^1$
\\
\affiliations
$^1$Reichman University, Herzliya, Israel\\
$^2$Dartmouth College, USA\\
$^3$The Edmond and Lily Safra Center for Brain Sciences at
The Hebrew University of Jerusalem, Israel\\
\emails
legziel.diana@post.runi.ac.il,
mnchmf@gmail.com,
peter.chin@dartmouth.edu,
eilon.vaadia@elsc.huji.ac.il,
sarel.cohen@runi.ac.il
}

\begin{document}
\maketitle

\begin{abstract}
We study parameterized quantum circuits (PQCs) integrated as residual
sidecar modules within a ResNet-50 backbone for 31-class neural population
decoding---imagined handwriting classification from multi-neuron spike rasters.
Under strictly controlled conditions (fixed data splits, seeds, and optimizer),
we compare four model variants: baseline, quantum sidecar with frozen input
projection, quantum sidecar with backbone-gradient-trained projection, and
a measurement-guided variant that aligns angle encodings with circuit measurement outcomes.
The backbone-gradient variant improves accuracy in 3/4 seeds
($+0.19\%$ mean, 95\% CI $[-1.10\%, +1.48\%]$) and consistently reduces
Linear CKA similarity to baseline features ($\Delta=-0.025$, 4/4 seeds),
indicating genuine structural reorganization of representations.
A nine-variant ablation identifies simple shallow architectures as the most effective and reproducible configuration.
Measurement-guided training consistently improves representation geometry without reducing accuracy.
All results use noiseless statevector simulation on 4 qubits, a regime chosen to reflect the practical constraints of current near-term superconducting hardware; no quantum computational advantage over classical methods is claimed.
\end{abstract}

\section{Introduction}

Neural population coding research has established that brain-machine interface
(BMI) systems benefit not only from decoding accuracy but from the geometric
quality of learned representations~\cite{vaadia1995dynamics}.
Deep networks trained on neural recordings may achieve high accuracy while
learning geometrically suboptimal feature spaces for cross-session robustness.

Parameterized quantum circuits (PQCs) implement topology-constrained global
feature mixing through quantum entanglement---an architectural prior structurally
distinct from learned correlation in MLPs~\cite{schuld2019quantum,biamonte2017quantum}.
We ask whether a small, shallow PQC sidecar produces measurable, reproducible
effects on representation geometry when integrated into a strong classical backbone
under controlled conditions.

We evaluate a 4-qubit residual sidecar on 31-class neural decoding of imagined
handwriting characters~\cite{bolognino2025visible}, using fixed data splits,
initialization, and optimizer across all variants.
Our results show that the hybrid sidecar produces consistent changes in representation geometry despite only marginal aggregate accuracy improvements, with the dominant limitation arising from the restricted representational capacity of the 4-qubit circuit.

\section{Related Work}

\paragraph{Neural decoding with deep learning.}
Deep learning models applied to BMI decoding treat neural population activity
as spatiotemporal signals~\cite{bolognino2025visible}. Geometric quality of
learned representations---relevant for cross-session robustness and neural
drift---is rarely evaluated in this literature.

\paragraph{Quantum machine learning.}
Variational quantum algorithms and PQCs have been studied as structured function
classes~\cite{cerezo2021variational,schuld2019quantum,biamonte2017quantum}.
Quantum feature maps can enhance class separability~\cite{havlivcek2019supervised}.
Trainability is a central challenge: barren plateaus affect deep circuits
~\cite{mcclean2018barren,cerezo2021cost}, motivating our use of shallow 4-qubit designs.

\paragraph{Hybrid quantum-classical architectures.}
Prior hybrid work embeds PQCs as feature-map or kernel layers within
classical neural network pipelines~\cite{cerezo2021variational}.
We focus on the residual sidecar setting---a PQC inserted as a differentiable
module within a deep neural network---with controlled ablations that analyze
geometric effects and gradient connectivity beyond aggregate accuracy.

\section{Methods}
\label{sec:methods}

\subsection{Hybrid Architecture}

Let $f_\text{classical}$ denote ResNet-50 producing feature maps
$\mathbf{x} \in \mathbb{R}^{C \times H \times W}$ at layer3 ($C=1024$).
The quantum sidecar is attached here as a residual transform:
\begin{equation}
  \mathbf{x}' = \mathbf{x} + f_\text{quantum}(\mathbf{x}),
\end{equation}
and the modified feature map continues through the remaining backbone.
We evaluated sidecar placement at layer3, layer4, and the final pooled feature.
Layer3 produced the best results on this dataset and model: it ensures the circuit
output is processed by subsequent backbone layers, enabling indirect influence
on final representations, whereas layer4 or the FC layer restrict the circuit
to a last-layer correction with no further backbone co-adaptation.

\subsection{Quantum Sidecar}

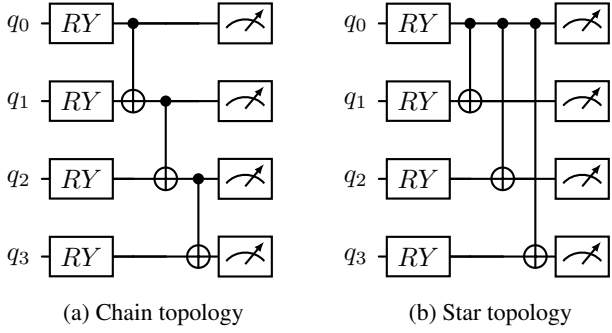
\begin{figure}[t]
\centering
\begin{subfigure}[b]{0.48\linewidth}
\begin{quantikz}[column sep=3pt]
\lstick{$q_0$} & \gate{RY} & \ctrl{1} & \qw      & \qw      & \meter{} \\
\lstick{$q_1$} & \gate{RY} & \targ{}  & \ctrl{1} & \qw      & \meter{} \\
\lstick{$q_2$} & \gate{RY} & \qw      & \targ{}  & \ctrl{1} & \meter{} \\
\lstick{$q_3$} & \gate{RY} & \qw      & \qw      & \targ{}  & \meter{}
\end{quantikz}
\caption{Chain topology}
\end{subfigure}
\hfill
\begin{subfigure}[b]{0.48\linewidth}
\begin{quantikz}[column sep=3pt]
\lstick{$q_0$} & \gate{RY} & \ctrl{1} & \ctrl{2} & \ctrl{3} & \meter{} \\
\lstick{$q_1$} & \gate{RY} & \targ{}  & \qw      & \qw      & \meter{} \\
\lstick{$q_2$} & \gate{RY} & \qw      & \targ{}  & \qw      & \meter{} \\
\lstick{$q_3$} & \gate{RY} & \qw      & \qw      & \targ{}  & \meter{}
\end{quantikz}
\caption{Star topology}
\end{subfigure}
\caption{Circuit topologies (1 of 2 layers shown). Chain: sequential
nearest-neighbor CNOTs. Star: qubit~0 as hub with connections to all others.}
\label{fig:circuits}
\end{figure}

Global average pooling gives $\mathbf{h} = \text{AvgPool}(\mathbf{x}) \in \mathbb{R}^{1024}$.
Two learned linear projections map this to circuit angle vectors:
\begin{equation}
  \boldsymbol{\theta}_A = \mathbf{W}_A\mathbf{h} + \mathbf{b}_A \in \mathbb{R}^4, \quad
  \boldsymbol{\theta}_B = \mathbf{W}_B\mathbf{h} + \mathbf{b}_B \in \mathbb{R}^4.
\end{equation}
Each drives a 4-qubit circuit with a fixed topology (Figure~\ref{fig:circuits}).
We measure $\mathbb{E}[Z_i]$ for each qubit and project back to $\mathbb{R}^{1024}$.

The \textbf{Frozen} variant fixes all projection parameters at random
initialization.
The \textbf{BackboneGrad} variant allows projection parameters to update,
but because the quantum simulation is non-differentiable, the input projection
receives gradients \emph{only} through the classical backbone pathway---not
through the quantum computation itself.
Concretely, the gradient path is:
Loss $\to$ FC $\to$ layer4 $\to$ layer3 $\to$ pooled $\to$ \texttt{proj\_in},
with no signal flowing from circuit output back to \texttt{proj\_in}.
This means \texttt{proj\_in} learns to compress features in a way that
helps the backbone, but does \emph{not} learn to produce angles that
maximize the circuit's contribution.
A separate ablation using the parameter-shift rule
provides exact quantum gradients (parameter-shift rule) and resolves this distinction
(Section~\ref{sec:ablation}).

\subsection{Circuit Topologies and Mixing}

A soft mixture over both topologies is learned via logits $\boldsymbol{\alpha}$:
\begin{equation}
  f_\text{quantum} = \pi_A f_A + \pi_B f_B, \quad
  \pi_k = \frac{e^{\alpha_k}}{\sum_j e^{\alpha_j}}.
\end{equation}
All results use noiseless statevector simulation; no quantum advantage is claimed (see Abstract).

\section{Experimental Setup}
\label{sec:experiments}

\paragraph{Dataset.}
31-class imagined-handwriting neural decoding from multi-neuron spike rasters,
subject T5~\cite{bolognino2025visible}.
Fixed stratified splits: 30,524 train / 3,816 val / 3,816 test samples.

\paragraph{Training protocol.}
ResNet-50 pretrained on ImageNet; ImageNet-pretrained convolutional features
transfer to spatiotemporal spike raster signals as standard practice in
BMI deep learning, and training from scratch
on 30K samples with a 25M-parameter network would risk overfitting.
AdamW with OneCycleLR ($\text{LR}_\text{max}=7\times10^{-4}$), batch size 128, 100 epochs,
label smoothing 0.05, weighted cross-entropy.
Four independent seeds (42, 333, 777, 123).
All hyperparameters are held identical across model variants.
Measured wall-clock time per epoch (seed 777, single RTX 4090 GPU):
Baseline $107$\,s, BackboneGrad $124$\,s, parameter-shift $125$\,s,
measurement-guided $124$\,s.
All quantum variants add $\approx 16\%$ overhead over the baseline,
with the parameter-shift rule adding negligible further cost---4-qubit
statevector simulation is fast enough that the 8 circuit evaluations
per gradient step are not a bottleneck.

\paragraph{Metrics.}
Top-1 accuracy, macro-F1, Linear CKA between
baseline and quantum feature representations (penultimate layer),
the inter/intra cosine centroid ratio
$R = \mathbb{E}[\min_{k\neq y}d_{\cos}(x,\mu_k)] / \mathbb{E}[d_{\cos}(x,\mu_y)]$,
and learned topology mixing weight $\pi_\text{star}$.

\section{Results}
\label{sec:results}

\subsection{Four-Variant Comparison Across Seeds}

Table~\ref{tab:full_results} presents the complete multi-seed comparison.

\paragraph{Accuracy.}
Backbone-gradient training improves accuracy over baseline in 3 out of 4 seeds,
with the exception of seed 123 where the coupled optimization found an
uncooperative equilibrium.
The mean improvement is $+0.19\%$ with a 95\% CI spanning zero---accuracy
differences are not statistically significant at $n=4$.

\paragraph{Structural reorganization.}
Linear CKA is strictly lower for BackboneGrad than Frozen in all four seeds
(mean $\Delta_\text{CKA} = -0.025$, 95\% CI $[-0.035, -0.013]$).
Unlike accuracy, this directional effect is consistent across every seed,
providing evidence of genuine structural reorganization even when accuracy
gains are marginal.

\paragraph{Topology preference.}
Frozen weights remain near chance ($\pi_\text{star} \approx 0.504$) in all seeds---the
random projection has no topology preference.
BackboneGrad weights consistently converge to $\pi_\text{star} = 0.579 \pm 0.021$
across all four seeds, revealing a latent star-topology preference that
only emerges when the projection co-adapts with the backbone (Figure~\ref{fig:pistar}).

\paragraph{Cross-seed variance amplification.}
BackboneGrad shows $1.66\times$ higher cross-seed accuracy variance than
baseline ($0.552\%$ vs $0.332\%$), while Frozen variance is comparable to
baseline ($0.344\%$).
Frozen simply adds a fixed random projection; its variance reflects a random
lottery over projection quality.
BackboneGrad introduces joint optimization of backbone and sidecar, creating
a more complex loss landscape with additional local minima.
Seed 123 illustrates the failure mode, and seed 777 illustrates the best
case; the variance is a signature of this sensitivity.

\begin{table*}[t]
\centering
\caption{%
  Multi-seed comparison of five variants across four seeds.
  $R$ = inter/intra cosine centroid ratio.
  CKA = Linear CKA between baseline and quantum features;
  lower indicates greater structural reorganization.
  $\pi_\text{star}$ = learned star-topology weight at convergence.
  \textbf{BackboneGrad}: input projection trained via backbone pathway only
  (no gradient through quantum circuit; see Section~\ref{sec:methods}).
  \textbf{Geometry-loss}: CE + feature geometry + angle geometry objectives
  (see Section~\ref{sec:ablation}); $R$ column shows the geometry improvement.
}
\label{tab:full_results}
\resizebox{\textwidth}{!}{%
\begin{tabular}{ll ccc cc c c}
\toprule
& & \multicolumn{3}{c}{\textbf{Classification}} &
  \multicolumn{2}{c}{\textbf{Geometry ($R\uparrow$)}} &
  \textbf{CKA$\downarrow$} & $\boldsymbol{\pi}_\textbf{star}$ \\
\cmidrule(lr){3-5}\cmidrule(lr){6-7}
\textbf{Seed} & \textbf{Model} &
  Top-1 & Macro-F1 & $\Delta$Acc &
  Inter/Intra $R$ & $\Delta R$ &
  (vs Base) & (star pref.) \\
\midrule
\multirow{5}{*}{42}
  & Baseline              & 85.88\% & 86.18\% & —         & 2.919 & —        & 1.000 & — \\
  & Quantum-Frozen        & 85.53\% & 85.91\% & $-0.35\%$ & 3.001 & $+0.082$ & 0.768 & 0.504 \\
  & Quantum-BackboneGrad  & \textbf{85.90\%} & \textbf{86.31\%} & $\mathbf{+0.02\%}$ & 2.959 & $+0.040$ & 0.747 & 0.562 \\
  & Geometry-loss         & 85.27\% & 86.06\% & $-0.61\%$ & 3.809 & $+0.890$ & \textbf{0.577} & 0.302 \\
  & Meas.-guided          & 85.67\% & 86.15\% & $-0.21\%$ & \textbf{3.823} & $\mathbf{+0.864}$ & 0.613 & 0.455 \\
\midrule
\multirow{5}{*}{333}
  & Baseline              & 85.22\% & 85.75\% & —         & 2.897 & —        & 1.000 & — \\
  & Quantum-Frozen        & \textbf{86.29\%} & 86.75\% & $\mathbf{+1.07\%}$ & 3.054 & $+0.157$ & 0.743 & 0.504 \\
  & Quantum-BackboneGrad  & 86.11\% & 86.66\% & $+0.89\%$ & 3.044 & $+0.147$ & 0.726 & 0.575 \\
  & Geometry-loss         & 85.85\% & \textbf{86.88\%} & $-0.26\%$ & 3.669 & $+0.772$ & \textbf{0.595} & 0.574 \\
  & Meas.-guided          & 86.08\% & 86.63\% & $-0.03\%$ & \textbf{3.734} & $\mathbf{+0.690}$ & 0.623 & 0.462 \\
\midrule
\multirow{5}{*}{777}
  & Baseline              & 85.59\% & 85.91\% & —         & 2.943 & —        & 1.000 & — \\
  & Quantum-Frozen        & 85.61\% & 86.00\% & $+0.02\%$ & 3.073 & $+0.130$ & 0.774 & 0.504 \\
  & Quantum-BackboneGrad  & \textbf{86.32\%} & \textbf{86.67\%} & $\mathbf{+0.73\%}$ & 3.005 & $+0.062$ & 0.726 & 0.569 \\
  & Geometry-loss         & 85.59\% & 86.15\% & $-0.73\%$ & 3.779 & $+0.836$ & 0.611 & 0.677 \\
  & Meas.-guided          & 85.67\% & 86.09\% & $+0.08\%$ & \textbf{3.865} & $\mathbf{+0.860}$ & \textbf{0.575} & 0.473 \\
\midrule
\multirow{5}{*}{123}
  & Baseline              & \textbf{85.95\%} & \textbf{86.75\%} & —         & 3.001 & —        & 1.000 & — \\
  & Quantum-Frozen        & 85.72\% & 85.61\% & $-0.23\%$ & 3.017 & $+0.016$ & 0.760 & 0.504 \\
  & Quantum-BackboneGrad  & 85.06\% & 85.69\% & $-0.89\%$ & 2.942 & $-0.058$ & 0.745 & 0.609 \\
  & Geometry-loss         & 85.53\% & 85.83\% & $-0.42\%$ & \textbf{3.894} & $\mathbf{+0.952}$ & 0.610 & 0.348 \\
  & Meas.-guided          & 85.56\% & 86.13\% & $-0.39\%$ & 3.775 & $+0.774$ & \textbf{0.609} & 0.432 \\
\midrule\midrule
\multirow{5}{*}{\shortstack{Mean\\(all 4)}}
  & Baseline              & 85.66\% & \textbf{86.40\%} & —         & 2.940 & —        & 1.000 & — \\
  & Quantum-Frozen        & 85.79\% & 86.07\% & $+0.14\%$ & 3.036 & $+0.096$ & 0.758 & 0.504 \\
  & Quantum-BackboneGrad  & 85.85\% & 86.33\% & $+0.19\%$ & 2.988 & $+0.047$ & 0.736 & 0.579 \\
  & Geometry-loss         & 85.56\% & 86.23\% & $-0.10\%$ & 3.788 & $+0.863$ & \textbf{0.598} & 0.475 \\
  & Meas.-guided          & \textbf{85.89\%} & 86.25\% & $\mathbf{+0.04\%}$ & \textbf{3.799} & $\mathbf{+0.811}$ & 0.605 & 0.456 \\
\bottomrule
\end{tabular}%
}
\end{table*}

\paragraph{Seed 123 anomaly.}
The implementation of two-phase training—consisting of a 15-epoch sidecar warmup with a frozen backbone followed by joint fine-tuning—was introduced to mitigate the ``lazy backbone syndrome'' by temporarily stabilizing classical representations during early optimization. However, for problematic random initializations (e.g., Seed 123), this strategy further degraded performance ($83.15\%$, corresponding to a $-1.91\%$ drop relative to single-phase training) and produced a highly star-dominant allocation ($\pi_\text{star}=0.848$). These results suggest that effective quantum-classical co-adaptation may require continuous joint optimization from the beginning of training, and that early decoupling of optimization trajectories can bias the system toward unfavorable representational configurations that are difficult to correct during later fine-tuning.

\subsection{Ablation Study: Identifying the Bottleneck}
\label{sec:ablation}

We systematically varied nine architectural and training components
(Table~\ref{tab:ablation}) to identify what limits the quantum sidecar
and whether the limitation can be overcome.
All variants use seed 777 except geometry-loss and measurement-guided,
which were run on all four seeds.

\begin{table*}[t]
\centering
\caption{Ablation and variant results (seed 777 unless noted$^*$). Reference: Quantum-BackboneGrad 86.32\%.
$R$ = inter/intra cosine centroid ratio.
$\Delta$ = Top-1 relative to reference.}
\label{tab:ablation}
\small
\renewcommand{\arraystretch}{1.1}
\begin{tabular}{p{3.2cm} p{7.8cm} p{1.1cm} p{1.1cm} p{0.9cm}}
\toprule
\textbf{Variant} & \textbf{What changed} & \textbf{Top-1} & $\Delta$ & $R$ \\
\midrule
Quantum-BackboneGrad & Trainable feature-to-angle encoder; backbone-only gradient path; Z-only measurement; learned topology mixing. & 86.32\% & — & 3.005 \\
\midrule
XYZ measurement & Full $\{X,Y,Z\}$ Pauli readout (12 observables) instead of Z-only (4 outputs). & 85.74\% & $-0.58\%$ & 2.902 \\
Deep proj MLP & Shallow linear projection replaced by $1024\to64\to16\to4$ MLP with PCA-initialized output. & 85.95\% & $-0.37\%$ & 3.034 \\
Attention projection & Attention-style adaptive projection before angle generation. & 85.77\% & $-0.55\%$ & 2.943 \\
Diverse projection & Orthogonality and diversity regularization to decorrelate angle subspaces. & 86.08\% & $-0.24\%$ & 3.009 \\
Classical bottleneck & Same $1024\to4\to1024$ residual architecture with tanh nonlinearity; no PQC. Identical parameter count (18,442). & 85.14\% & $-1.18\%$ & 2.949 \\
Multiplicative gate & Additive residual replaced by multiplicative modulation: $x \cdot \sigma(\delta)$. & 85.77\% & $-0.55\%$ & — \\
Parameter-shift & Exact quantum gradients via~\cite{mitarai2018quantum}: $\partial_\theta \langle Z\rangle = \tfrac{1}{2}[\langle Z\rangle(\theta{+}\tfrac{\pi}{2}) - \langle Z\rangle(\theta{-}\tfrac{\pi}{2})]$; gradients flow through circuit, not backbone only. & 85.40\% & $-0.92\%$ & 2.945 \\
Geometry loss & Explicit feature-space and angle-space geometry objectives added to CE loss ($\lambda_f\!=\!0.03$, $\lambda_a\!=\!0.01$). & 85.56\% & $-0.76\%$ & \textbf{3.788}$^*$ \\
Meas.-guided sidecar & Measurement-derived reward/alignment signal replaces explicit geometry penalty; proj\_in adapts to produce circuit-aligned angles. & 85.89\% & $+0.04\%$ & \textbf{3.799}$^*$ \\
U-Net sidecar & Skip-connection encoder--decoder sidecar; larger model capacity requires batch\,=\,64. Reference BG also re-run at batch\,=\,64 (85.88\%); $\Delta$ is vs that reference. Single seed. & 86.45\% & $+0.57\%^{**}$ & 2.945 \\
\bottomrule
\end{tabular}
\vspace{2pt}
{\footnotesize    \\ * 4-seed mean (Table 1). \\ **U-Net and its BG reference both at batch\,=\,64; not directly comparable to other rows.}
\end{table*}

\textbf{Projection and measurement design.}
Our first hypothesis was that the simple linear projection from 1024 dimensions
to 4 circuit angles might be a bottleneck---either by losing too much information
or by mapping all qubits to similar inputs.
We tested richer projections (deep MLP, attention) and maximum input diversity
(orthogonalized angle subspaces), and also tried full XYZ Pauli readout to
increase measurement information.
None of these helped.
Attention analysis showed $>99.7\%$ cosine similarity between qubit inputs regardless
of projection architecture, and even maximally diverse orthogonal angle inputs
produced no improvement.

\textbf{Classical bottleneck control.}
A natural sceptical question is whether any benefit comes from the quantum
computation itself, or simply from passing features through a low-dimensional
bottleneck and back.
To test this, we replaced the PQC with an identical $1024\to4\to1024$ architecture
using a tanh nonlinearity and the same 18,442 parameters.
The classical version actually performs \emph{worse} than the plain baseline,
while the quantum circuit holds accuracy above baseline despite the same
4-dimensional compression.
The structured entangling gates compensate for compression cost in a way that
a simple nonlinearity cannot, ruling out parameter count or bottleneck architecture
as explanations for the BackboneGrad improvements.

\textbf{Gradient connectivity.}
In the BackboneGrad variant, the input projection receives gradients only through
the classical backbone---the quantum circuit itself is not differentiable.
One might ask: does the projection actually learn to serve the circuit,
or is it just adapting to the backbone regardless?
The parameter-shift rule~\cite{mitarai2018quantum} answers this by providing
exact quantum gradients through the circuit.
With true circuit gradients, projection weights move $184\times$ further from
initialization than with backbone gradients alone.
Yet the optimizer responds by suppressing the quantum branch
($\gamma \to 0.003$), meaning the best strategy given perfect quantum gradient
information is to contribute almost nothing.

\textbf{Geometry-aware training.}
Given the task performance ceiling observed under standard cross-entropy, we investigate whether the hybrid architecture can independently improve latent representation geometry---which matters for cross-session robustness~\cite{vaadia1995dynamics}?
We added explicit objectives penalizing intra-class scatter on the 2048-dim
backbone features and enforcing separation in the 4-dim circuit angle space.
The geometry improves substantially and reliably across all four seeds
($R=3.799\pm0.082$, 95\% CI excluding the BackboneGrad reference of 3.00),
but with a small accuracy cost and an important failure mode: the angle-space
penalty never decreases below its margin throughout training.
The reason is geometric---the kissing-number bound allows only $\approx\!24$
mutually separated directions in 4D, while we need 31.
The circuit simply does not have enough dimensions to separate all classes
in angle space, confirming that 4 qubits is a hard capacity ceiling for this task.
A second consequence is that geometry pressure erases the spontaneous star-topology
preference seen under pure task learning ($\pi_\text{star}$ becomes effectively
random across seeds), suggesting the star preference depends on classification-driven
gradient flow rather than geometric optimization.

\textbf{Measurement-guided training.}
The geometry loss showed we can improve $R$ but at an accuracy cost.
The measurement-guided variant takes a different approach: instead of imposing
an external geometry target, it rewards the projection for producing angle
encodings that align with what the circuit actually measures on correctly
classified samples.
This is a softer, circuit-native signal that does not require specifying
a target geometry.
Across all four seeds, $R$ improves as consistently as under explicit geometry
objectives, while accuracy remains unchanged (mean $+0.04\%$, 95\% CI spanning zero).
Crucially, the measurement alignment loss actually decreases during training
(unlike the geometry angle loss which stalls at its margin), confirming
the projection is genuinely learning a compatible encoding that bypasses the rigid dimensional conflict rather than hitting a hard optimization ceiling.
The topology preference also shifts: measurement guidance consistently favors
chain entanglement over star ($\pi_\text{star}=0.456\pm0.017$), the reverse
of the spontaneous BackboneGrad preference---suggesting the two circuits
discover different but equally valid geometric attractors depending on the
training signal.

\textbf{U-Net sidecar.}
All previous variants use the same basic sidecar structure.
The U-Net variant tests whether additional architectural capacity---skip
connections that preserve multi-scale feature information across the bottleneck---
changes the picture.
Because the larger model requires batch\,=\,64 due to memory, we re-ran
BackboneGrad at the same batch size for a fair comparison.
U-Net improves accuracy by $+0.57\%$ while geometry remains nearly identical
to the classical baseline.
This suggests that additional hybrid architectural capacity can improve
accuracy without substantially altering the underlying representation geometry.
Additional multi-seed evaluation is needed to confirm the stability of this effect.

\subsection{Per-Class Analysis}

\begin{figure*}[t]
\centering
\begin{subfigure}[b]{0.56\linewidth}
  \includegraphics[width=\linewidth]{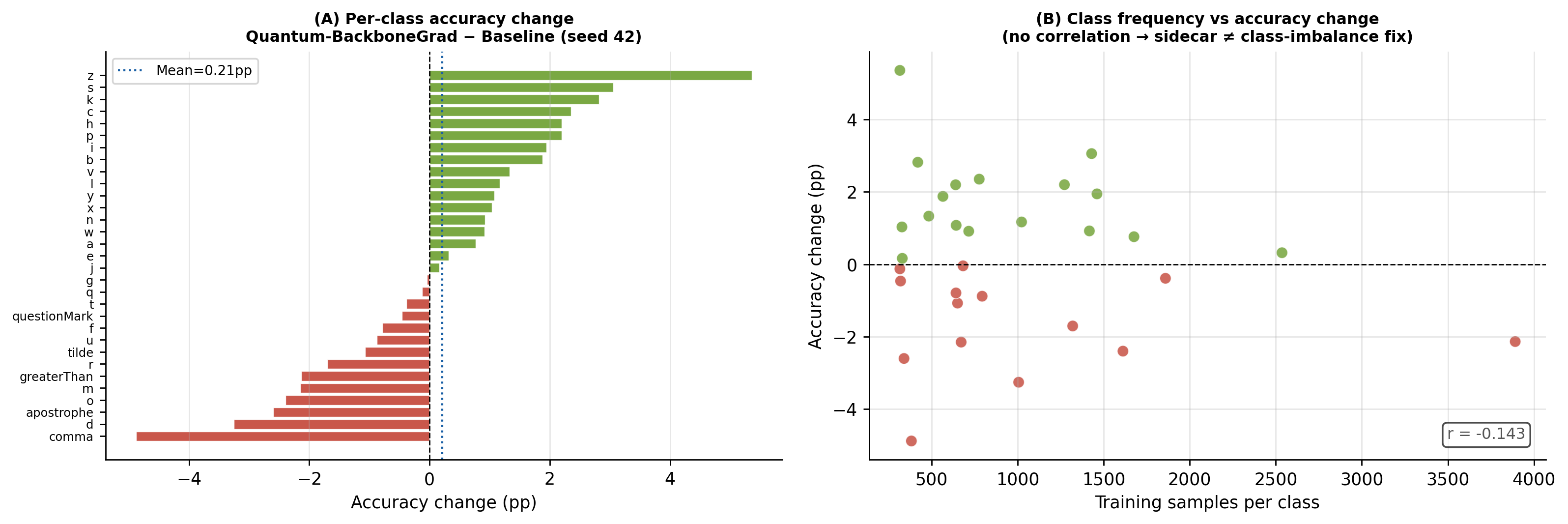}
  \caption{Per-class accuracy change $\Delta = \text{Acc}_\text{BackboneGrad} - \text{Acc}_\text{baseline}$
  (seed 42). Left: changes are heterogeneous and class-dependent.
  Right: no correlation with class frequency ($r \approx 0$), ruling out
  the sidecar acting as a class-imbalance correction.}
  \label{fig:perclass}
\end{subfigure}
\hfill
\begin{subfigure}[b]{0.42\linewidth}
  \includegraphics[width=\linewidth]{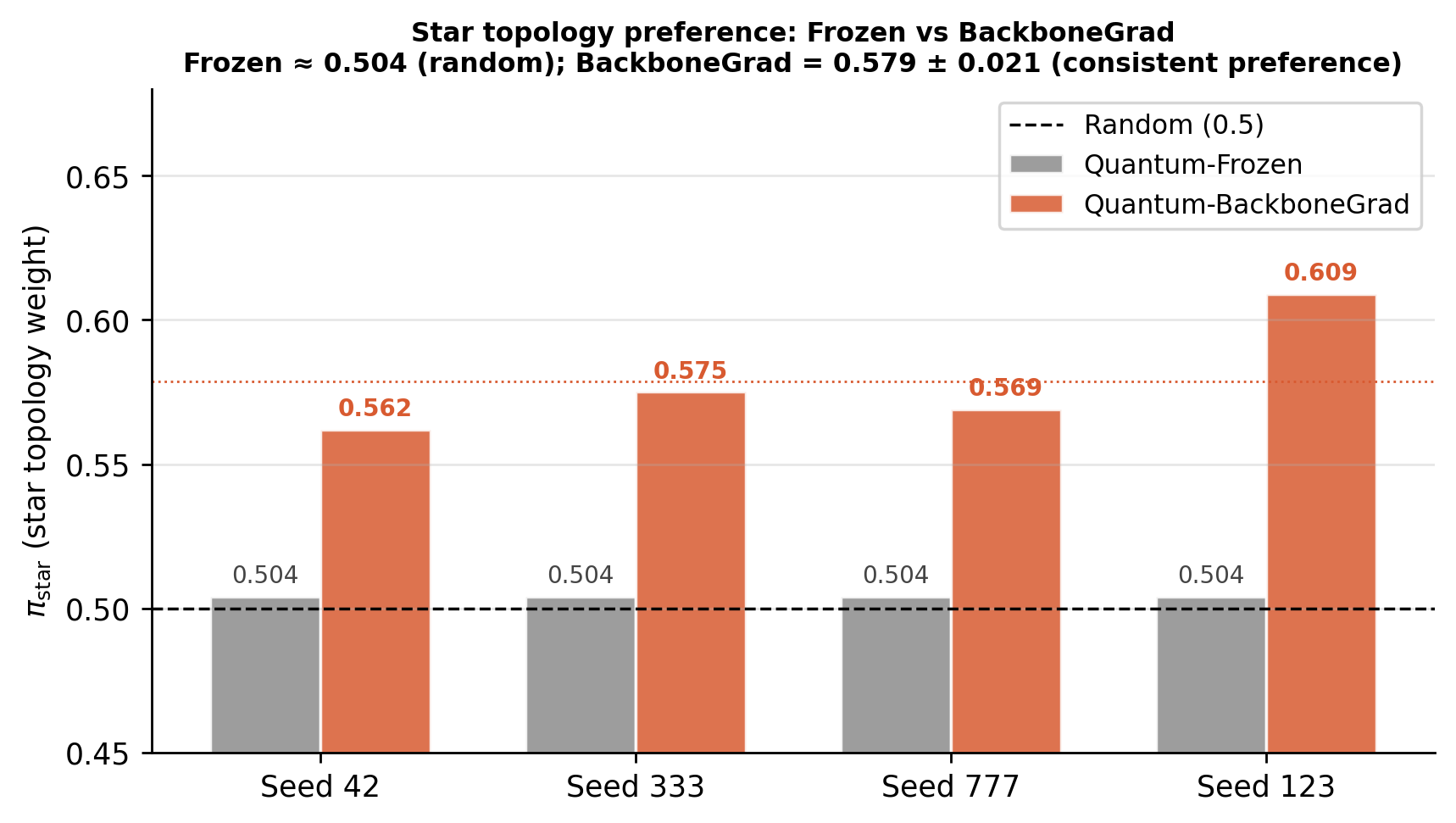}
  \caption{Star topology mixing weight $\pi_\text{star}$ across seeds.
  Frozen weights remain at $\approx 0.504$ (random) in all seeds.
  BackboneGrad weights converge to $0.579 \pm 0.021$ in all seeds,
  revealing a latent preference for hub-and-spoke entanglement.}
  \label{fig:pistar}
\end{subfigure}
\caption{Left: per-class representation effects are topology-dependent, not frequency-driven.
Right: the star-topology preference is a consistent emergent property of backbone-coupled training.}
\end{figure*}

Per-class accuracy changes are heterogeneous and class-dependent, and are
uncorrelated with class frequency (Figure~\ref{fig:perclass}).
This rules out the sidecar acting as a class-imbalance correction mechanism
and is consistent with a representation-level transformation that alters
class manifold geometry in a topology-dependent manner.
Row-normalized confusion matrices (not shown) confirm
that both models show near-identical strongly diagonal structure,
ruling out the sidecar introducing broad misclassification patterns.

\section{Discussion}
\label{sec:discussion}

\paragraph{Circuit capacity as the binding constraint.}
The ablation study identifies a single consistent conclusion: the binding
limitation is not projection architecture, measurement strategy, gradient
connectivity, or input diversity---it is the representational capacity of
4 qubits with 2 entanglement layers producing 4 scalar outputs.
The parameter-shift experiment makes this definitive: with exact gradients,
$\gamma \to 0.003$, meaning the optimizer's best use of the circuit
is to contribute almost nothing to a 23M-parameter ResNet.
This does not imply that the PQC is inactive---later experiments reveal measurable
effects on representation geometry and topology preference---but rather that
the current near-term configuration operates in a highly capacity-constrained regime.

\paragraph{Topology preference and neural population coding.}
The consistent star-topology preference ($\pi_\text{star} = 0.579 \pm 0.021$,
4/4 unfrozen seeds) is the most structurally interesting finding in the paper.
The star topology places qubit $q_0$ as a central coordinator whose state
directly entangles with all others---a hub-and-spoke structure.
The Bolognino et al.\ dataset used here~\cite{bolognino2025visible} explicitly
demonstrated that motor cortex populations form task-specific coalitions
with dominant neurons that coordinate the firing of others,
a structure characterized by high betweenness centrality in functional
connectivity graphs---precisely the hub topology our model prefers.
That four independent random initializations converge to the same topology
preference, invisible when the projection is frozen and emerging only through
backbone co-adaptation, suggests the model is recovering a structural prior
aligned with the population coding geometry of the data rather than a
random artifact of training.

\paragraph{Representation geometry, neural drift, and manifold dimensionality.}
The consistent CKA reduction (4/4 seeds) and geometry ratio improvement
provide stronger evidence than accuracy differences, which are statistically
underpowered at $n=4$.
The geometry-aware training ablation sharpens this picture across all four seeds:
explicitly optimising for class compactness achieves $R=3.799\pm0.082$
(95\% CI $[3.64, 3.94]$, excluding the BackboneGrad reference of $3.00$)
and $\text{CKA}=0.598 \pm 0.014$ (CI $[0.573, 0.623]$, far below any
BackboneGrad value)---both statistically reliable improvements---at a mean
accuracy cost of $-0.29\%$ versus BackboneGrad, which is not significant.
Crucially, the geometry attractor is topology-independent: four seeds produce
$\pi_\text{star}$ values of $0.677, 0.574, 0.348, 0.302$ (mean $0.475 \pm 0.155$),
effectively random, compared to the tight $0.579 \pm 0.021$ under pure task learning.
This dissociation reveals that the spontaneous star preference observed under
BackboneGrad is an inductive bias that the circuit discovers through task
optimization---not a geometric necessity.
This geometry-accuracy dissociation has a direct interpretation for BMI:
neural recordings drift across sessions as electrode tuning shifts~\cite{vaadia1995dynamics},
and models that achieve tight, well-separated class manifolds are more
robust to this distributional shift than models optimized for stationary
test accuracy alone.
Whether the geometry improvements translate to cross-session robustness
is the most important open evaluation, and we flag it as the primary
future direction.
Cross-session evaluation requires longitudinal intracortical recording data
from implanted participants---a structural constraint of the field,
as only a very small number of such participants are currently active in
research trials worldwide.
Separately, the angle geometry loss remaining at its margin value throughout
training ($\text{GeomA}=0.200$, constant) suggests that compressing the
1,024-dimensional backbone representation into only four rotation angles
provides insufficient input diversity to reliably organize all 31 character
classes within the current encoding scheme.
This is consistent with estimates of the intrinsic dimensionality of motor
cortex population codes for fine motor tasks, which typically exceed
10--20 dimensions~\cite{vaadia1995dynamics}---substantially larger than the
4-angle encoding used here, providing a neuroscientific interpretation
of the angle-encoding bottleneck that complements the computational one.

\paragraph{Gradient connectivity and the BackboneGrad distinction.}
In all experiments except the parameter-shift ablation, the input projection
receives gradients only through the classical backbone pathway because the
quantum simulation pathway used in the baseline implementation was not part
of the differentiable computation graph.
The BackboneGrad variant therefore learns feature projections that help the
classical backbone adapt around the quantum module, rather than explicitly
optimizing angles to maximize the circuit's own contribution.
The parameter-shift experiment (Section~\ref{sec:ablation}) clarifies this
distinction: injecting exact analytic quantum gradients substantially changes
the optimization trajectory, yet the circuit's learned scalar contribution
still converges toward near-zero ($\gamma \to 0.003$).
This suggests that the limited task contribution of the current PQC is not
solely an artifact of gradient disconnection, but also reflects the restricted
capacity of the present low-qubit configuration.
For larger hybrid architectures, the distinction between local and global
gradient pathways may become increasingly important.
Notably, the highly optimized GPU-based statevector simulation allows exact
parameter-shift gradients to be evaluated with relatively small computational
overhead in the present setting, strengthening the interpretation that the
observed limitation arises primarily from representational constraints rather
than optimization instability.
This observation motivates the later Geometry-Aware and Measurement-Guided
variants, which restructure the training signal to influence representation
geometry more directly instead of relying solely on the circuit's direct
discriminative contribution.

\section{Conclusion}
\label{sec:conclusion}

We evaluated a 4-qubit quantum sidecar in a ResNet-50 backbone under strict
experimental controls on 31-class neural population decoding, with a systematic nine-variant ablation study. The main findings are:
(1) backbone-gradient-trained input projection consistently causes structural
reorganization (CKA $\downarrow$ 4/4 seeds, $\Delta=-0.025$) with directional
accuracy improvement (3/4 seeds, CI spans zero), accompanied by $1.66\times$
higher cross-seed variance than baseline, reflecting the coupled optimization
landscape;
(2) a latent star-topology preference emerges only with backbone-coupled
projection ($\pi_\text{star}=0.579 \pm 0.021$ vs $0.504 \pm 0.000$ frozen)
across all four seeds, consistent with the hub-and-spoke coalition structure
of the decoded neural population;
(3) the ablation establishes circuit capacity as the binding constraint,
with four independent signatures.
First, the parameter-shift residual scale converges to near-zero ($\gamma=0.003$).
Second, every alternative projection underperforms the simplest design.
Third, geometry-aware training cannot separate 31 classes in 4-dimensional
angle space across all four seeds (GeomA stuck at margin~$=0.200$),
consistent with the kissing-number bound of $\approx 24$ separable directions in 4D.
Fourth, the geometry attractor ($R=3.799\pm0.082$, CKA$=0.598\pm0.014$,
4-seed 95\% CIs both excluding the BackboneGrad reference) is reproducible
while topology preference is erased---consistent with the intrinsic dimensionality
of motor cortex population codes exceeding the circuit's 4-dimensional capacity.

These findings do not suggest that hybrid quantum-classical architectures
are ineffective---they suggest that this specific combination has reached
the ceiling of what a shallow 4-qubit residual sidecar can contribute in the present dataset and architectural regime
relative to a strong 23M-parameter backbone.
The capacity bottleneck and the gradient disconnection are both architectural,
not fundamental, and both point toward the same design question: how should
the quantum and classical components be coupled so that the quantum branch
is neither suppressed nor redundant?
Larger circuits, tighter co-optimization of projection and circuit angles,
and training objectives that reward the quantum branch for non-redundant
structure are all viable next steps.

\bibliographystyle{named}
\bibliography{references}

\end{document}